\documentclass{article}
\usepackage{spconf,amsmath,epsfig,graphicx,hyperref,microtype,booktabs,multirow}
\let\OLDthebibliography\thebibliography
\renewcommand\thebibliography[1]{
  \OLDthebibliography{#1}
  \setlength{\parskip}{0pt}
  \setlength{\itemsep}{0pt plus 0.3ex}
}
\usepackage{amsmath,amsfonts,bm}

\def\eqref#1{equation~\ref{#1}}

\def\1{\bm{1}}

\DeclareMathAlphabet{\mathsfit}{\encodingdefault}{\sfdefault}{m}{sl}
\SetMathAlphabet{\mathsfit}{bold}{\encodingdefault}{\sfdefault}{bx}{n}

\usepackage{fancyhdr}
\begin{document}\sloppy

\title{Enhancing Scientific Figure Captioning Through Cross-modal Learning}

\name{Mateo Alejandro Rojas, Rafael Carranza}
\address{Technological University of Peru}

\maketitle

\begin{abstract}
Scientific charts are essential tools for effectively communicating research findings, serving as a vital medium for conveying information and revealing data patterns. With the rapid advancement of science and technology, coupled with the advent of the big data era, the volume and diversity of scientific research data have surged, leading to an increase in the number and variety of charts. This trend presents new challenges for researchers, particularly in efficiently and accurately generating appropriate titles for these charts to better convey their information and results.
Automatically generated chart titles can enhance information retrieval systems by providing precise data for detailed chart classification. As research in image captioning and text summarization matures, the automatic generation of scientific chart titles has gained significant attention. By leveraging natural language processing, machine learning, and multimodal techniques, it is possible to automatically extract key information from charts and generate accurate, concise titles that better serve the needs of researchers. This paper presents a novel approach to scientific chart title generation, demonstrating its effectiveness in improving the clarity and accessibility of research data.
\end{abstract}


\section{Introduction}
\label{sec:intro}

The rapid advancement of deep learning technology and the burgeoning era of big data have brought multimodal data processing to the forefront of current research trends \cite{Zhang2023,zhou2024visual,zhou2022claret}. Scientific charts, which combine image and text modalities, have become indispensable tools for researchers to intuitively convey complex data patterns and insights. These charts play a pivotal role in various fields by visually representing data relationships and features, and providing textual descriptions to complement visual information \cite{Baltrusaitis2018}.

Despite their importance, the automatic generation of titles for scientific charts has not been fully explored, and current research primarily focuses on unimodal methods. These methods often fail to capture the complete essence of the charts, leading to titles that may not accurately reflect the core insights. This limitation is particularly evident when the data is unstable, resulting in significant drops in the quality of generated titles from model \cite{Arevalo2017,Li2018}.

Scientific charts integrate two types of information modalities: image and text. The image modality visually represents the relationships between chart elements and data features through visual components such as color and lines. However, images alone may be insufficient in clearly expressing specific content. Conversely, the text modality contains detailed descriptions of chart objects but has limitations in revealing the intrinsic connections between chart elements and conveying deep insights \cite{Li2018}.

The significance of the task of automatic scientific chart title generation lies in its potential to enhance the clarity and accessibility of research findings. Accurate and concise titles enable better comprehension and quicker assimilation of chart contents, which is critical in the fast-paced research environment. Furthermore, automated chart titles can facilitate the development of chart retrieval systems, aiding in the efficient organization and classification of vast amounts of research data \cite{Ramachandram2017}.

However, several challenges arise in this task. First, the heterogeneity of the data, where visual and textual elements must be effectively integrated, poses a significant hurdle \cite{Baltrusaitis2018}. Second, the variability in chart formats and the quality of data from different sources can affect the robustness of title generation models \cite{Mogadala2019}. Third, ensuring that the generated titles are not only syntactically correct but also semantically meaningful and contextually appropriate requires sophisticated modeling techniques \cite{Gao2020}.

Motivated by these challenges, this paper proposes a novel approach for generating scientific chart titles by leveraging deep learning techniques that utilize multimodal data. Our method aims to address the limitations of unimodal approaches by integrating both visual and textual information to generate more accurate and informative titles. The core of our approach involves using pre-trained models to encode chart images and text, followed by a co-self-attention mechanism to fuse these modalities into a cohesive feature representation.

Specifically, our proposed method works as follows:
\begin{enumerate}
    \item \textbf{Visual Encoding:} A pre-trained model, such as CLIP, is used to encode the chart images, capturing key visual features and structural information \cite{Radford2021,zhou2021improving,zhou2021modeling,wang2024memorymamba}.
    \item \textbf{Text Encoding:} The textual content within the charts is encoded using a large-scale text pre-trained language model like RoBERTa, to obtain rich contextual representations \cite{Liu2023SciCap,zhou2022eventbert}.
    \item \textbf{Modal Interaction:} The encoded visual and textual features are integrated using a co-self-attention mechanism, creating a multimodal feature representation that leverages both modalities \cite{Xu2015ShowAT}.
    \item \textbf{Title Generation:} The fused multimodal feature representation is then used to generate the title of the chart, utilizing advanced decoding techniques \cite{You2016ImageCW}.
\end{enumerate}

Experiments conducted on the SCICAP dataset demonstrate the effectiveness of our approach, showing significant improvements over baseline models. Our model's ability to utilize multimodal information results in titles that are not only accurate but also contextually relevant and meaningful.

\begin{itemize}
    \item Our method effectively combines visual and textual modalities using a co-self-attention mechanism, enhancing the quality and relevance of generated chart titles.
    \item By incorporating contrastive learning and relevant domain knowledge, our model improves its generalization and reasoning capabilities, leading to more robust performance across diverse datasets.
    \item We demonstrate the practical application of our method through the development of a scientific chart title generation system, significantly outperforming existing baseline models on the SCICAP dataset.
\end{itemize}

This paper is structured as follows: Section 2 reviews related work, Section 3 and Section 4 detail our proposed dataset and method, Section 5 presents our experimental setup and results, and Section 6 concludes with future research directions.

\section{Related Work}
\label{sec:related_work}

This section reviews the existing literature related to image captioning and figure captioning, highlighting the advancements and methodologies that have been proposed in these areas.

\subsection{Image Captioning}
Image captioning has been an active area of research, focusing on generating descriptive sentences for images. Early approaches relied on template-based methods and retrieval-based techniques. With the advent of deep learning, generative models have become the standard.

\textbf{Show and Tell} by Vinyals et al. introduced a generative model based on deep recurrent networks, significantly improving the quality of generated captions by learning a direct mapping from images to sentences using an encoder-decoder framework \cite{Vinyals2015ShowAT}. Building on this, \textbf{Show, Attend and Tell} by Xu et al. integrated an attention mechanism, allowing the model to focus on specific parts of an image while generating each word in the caption \cite{Xu2015ShowAT}.

\textbf{ClipCap} by Mokady et al. leveraged the CLIP model for encoding images, followed by a mapping network to generate captions compatible with language models \cite{Mokady2021ClipCap}. Other significant works include using large-scale pre-trained models like BERT and GPT for text generation, further enhancing the quality and coherence of the generated captions \cite{Radford2021LearningTV}.

\textbf{Semantic Attention} introduced by You et al. emphasized the importance of semantic attributes in image captioning, proposing an attention mechanism that focuses on meaningful parts of the image \cite{You2016ImageCW}. More recently, \textbf{CapText} by Zhang et al. explored the use of large language models for integrating both image context and textual descriptions, highlighting the potential of combining visual and textual modalities \cite{Zhang2023CapText}.

\subsection{Figure Captioning}
Figure captioning, a specialized branch of image captioning, focuses on generating descriptive texts for figures in scientific documents. This task involves unique challenges due to the complex and varied nature of scientific figures.

\textbf{Figure Captioning with Reasoning and Sequence-Level Training} by Wang et al. introduced a model that incorporates reasoning capabilities and sequence-level training to generate detailed and accurate figure captions \cite{Wang2019FigureCR}. Another innovative approach is presented in \textbf{Summaries as Captions} by Jin et al., where figure captioning is treated as a text summarization task, leveraging automated summarization models like PEGASUS \cite{Jin2023SummariesAC}.

The \textbf{SciCap+} dataset and associated model, described by Liu et al., augment figure captioning with contextual knowledge from scientific documents, significantly improving the relevance and accuracy of the generated captions \cite{Liu2023SciCap}. Similarly, \textbf{FigCaps-HF} by Lee et al. introduced a framework that incorporates human feedback to train models for higher-quality figure captions \cite{Lee2023FigCapsHF}.

The \textbf{ICCV 2023 Scientific Figure Captioning Challenge} highlighted the use of contrastive learning principles to enhance OCR-based image extraction and caption generation, showcasing the integration of multiple advanced techniques to address the complexities of scientific figure captioning \cite{ICCV2023}. Furthermore, \textbf{SciCapenter} by Zhang et al. supports caption composition for scientific figures by providing machine-generated suggestions and quality ratings \cite{Zhang2023SciCapenter}.

\section{SciCap v2}

We conducted our experiments using the SCICAP dataset. During the preparatory phase, a thorough analysis revealed significant quality variations due to the dataset's origin from the arXiv preprint platform. These variations raised concerns about the validity of some data, particularly due to the lack of necessary textual information in many charts, with some even missing basic data indicators.

To address these issues, we undertook extensive data cleaning. This involved removing records that lacked sufficient textual chart data, specifically those with text lengths of less than one symbol. After cleaning the data, we re-ran the baseline experiments initially conducted with the SCICAP dataset.

The results demonstrated a substantial improvement in experimental outcomes, despite the reduction in dataset size. The data cleaning process notably enhanced the overall quality and effectiveness of the dataset. The table below presents the BLEU-4 scores for different methods before and after data cleaning:

\begin{table}[t]  
  \centering  
  \begin{tabular}{ccc}  
    \toprule  
    \textbf{Datasets} & \textbf{Method} & \textbf{BLEU-4 Score} \\ \midrule 
    Original SciCap & Vision Only & 0.0220  \\
    & Text Only & 0.0200  \\ 
    & Vision+Text & 0.0210 \\ \midrule  
    SciCap v2 & Vision Only & 0.0380  \\
    & Text Only & 0.0970  \\ 
    & Vision+Text & 0.1230 \\ \bottomrule
  \end{tabular}  
  \caption{BLEU-4 Scores for Different Datasets and Methods}  
\end{table}

These results indicate that the cleaned dataset (SciCap v2) significantly outperformed the original dataset across all methods, affirming the positive impact of data cleaning on dataset quality and experimental effectiveness.

\section{Method}

\subsection{Problem Statement}
Given a dataset of paired images, texts, abstracts, and captions $\left\{x^{i}, t^{i}, a^{i}, c^{i}\right\}_{i = 1}^{N}$, our objective is to generate meaningful captions for unseen scientific chart images. Captions are represented as a series of tokens $c^{i} = c_{1}^{i}, \ldots, c_{\ell}^{i}$, with a predefined maximum length. The training objective is formulated as follows:
\begin{align}
\max _{\theta} \sum_{i=1}^{N} \log p_{\theta}\left(c_{1}^{i}, \ldots, c_{\ell}^{i} \mid x^{i}, t^{i}, a^{i}\right)
\end{align}
where $\theta$ denotes the trainable parameters of the model. Our approach combines encoded visual and textual information from the chart and integrates related abstract information as prior knowledge. Inspired by recent work, we consider conditions as prefixes of titles, using an auto-regressive language model to predict the next token. Thus, our goal can be described as:
\begin{align}
\max _{\theta} \sum_{i=1}^{N} \sum_{j=1}^{\ell} \log p_{\theta}\left(c_{j}^{i}, t_{j}^{i}, a_{j}^{i} \mid x^{i}, c_{1}^{i}, \ldots, c_{j-1}^{i}\right)
\end{align}

\subsection{Visual Encoding}
We extract visual features from images $x^{i}$ using the visual encoder of a pre-trained CLIP model. The CLIP embeddings are then mapped to $k$ embedding vectors using a multi-layer perceptron (MLP) network $M$:
\begin{align}
\mathbf{P} = p_{1}^{i}, \ldots, p_{k}^{i} = M\left(\operatorname{CLIP}\left(x^{i}\right)\right)
\end{align}

\subsection{Text Encoding}
Textual information is encoded using the RoBERTa language model. RoBERTa's tokenizer projects the text into a series of embeddings. This approach resolves the issue of encoding out-of-vocabulary words in the baseline model. The text is tokenized into a sequence of words $\left\{w_{0}, w_{1}, \ldots, w_{n}\right\}$ and encoded as a sequence of word embeddings $\left\{t_{0}, t_{1}, \ldots, t_{n}\right\}$:
\begin{align}
\mathbf{T} = t_{i} = \text{word-embedding}\left(w_{i}\right)
\end{align}
A normalization layer is applied before providing the word embedding sequence as input to the model.

\subsection{Incorporating Related Knowledge}
Abstract information related to scientific charts is encoded via a text encoder and feature compression. This relevant knowledge enhances the model's ability to accurately capture the core information and themes of the chart. When chart text information is insufficient, relevant knowledge supplements the text modality, enriching the generated titles. The RoBERTa model encodes the abstract information, which is not directly used but undergoes feature extraction. The feature vectors from RoBERTa word embeddings are input into a fully connected layer with an activation function, converting them into 2048-dimensional vectors. These encoded feature vectors are fused with the model's embedding vectors, forming a representation similar to template addition:
\begin{align}
\left[\mathrm{emb}[cls] + \operatorname{relu}\left(\mathrm{Mlp}\left(\mathrm{emb}\left[T_{\text{abstract}}\right]\right)\right) + \mathrm{emb}[cls] + M\right]
\end{align}
This method draws inspiration from template-based generative question answering, retaining the original embedding information while incorporating high-dimensional feature representations of relevant knowledge. The fused representation is then fed into a text decoder.

\subsection{Modal Interaction}
Modal interaction is achieved through a co-Transformer architecture, utilizing a co-attention mechanism. After processing by the text encoder, chart and text modal data pass through a fully connected layer, mapping the output to a 2048-dimensional feature vector. Positional encoding is added to the feature vectors before inputting them into the modal interaction module.

The multimodal fusion interaction module comprises two Transformer Blocks, each containing multiple Transformer Layers. Visual and textual modalities interact through a self-attention mechanism. Text and image features are mapped into 1024-dimensional vectors for attention computation. The result is residually connected with the input vector and undergoes layer normalization. The modal features are updated through multiple Transformer Layers, and the final outputs from different modal data Transformer Blocks are concatenated into a 2048-dimensional feature vector. The multimodal feature representation process can be summarized as:
\begin{align}
E = \text{co-Transformer}(\operatorname{Mlp}(H), C)
\end{align}

\subsection{Decoding}
An attention-based model serves as the decoder. The encoded visual features $\mathbf{P}$ and textual information $\mathbf{T}$ initialize the LSTM:
\begin{align}
\boldsymbol{c}_{0} &= \sigma\left(\left[\boldsymbol{W}_{P,Ic} \boldsymbol{P}, \boldsymbol{W}_{T,Ic} \boldsymbol{T}\right]\right) \\
\boldsymbol{h}_{0} &= \sigma\left(\left[\boldsymbol{W}_{P,Ih} \boldsymbol{P}, \boldsymbol{W}_{T,Ih} \boldsymbol{T}\right]\right)
\end{align}
where $\sigma(.)$ is the sigmoid function. Captions are encoded into contextual word vectors by the GPT-2 language model, preprocessed with BOS and EOS tokens. The contextual word vector $\mathbf{C}_{y, t}$ represents the word $y_{t}$ and is further embedded by a linear embedding $E$:
\begin{align}
\boldsymbol{e}_{0} &= \mathbf{0}, \text{ otherwise} \\
\boldsymbol{e}_{t} &= e\left(y_{t}\right) = \boldsymbol{E} \mathbf{C}_{\boldsymbol{y}, t}, t>0
\end{align}
The word vector $e_{t}$ and context vector $d_{t}$ are used as LSTM input. Signals for the input gate, forget gate, and output gate are:
\begin{align}
i_{t} &= \sigma\left(\boldsymbol{W}_{iy} e_{t} + \boldsymbol{W}_{ih} h_{t-1} + \boldsymbol{W}_{id} d_{t} + b_{i}\right) \\
f_{t} &= \sigma\left(\boldsymbol{W}_{fy} e_{t} + \boldsymbol{W}_{fh} h_{t-1} + \boldsymbol{W}_{fd} d_{t} + b_{f}\right) \\
\boldsymbol{o}_{t} &= \sigma\left(\boldsymbol{W}_{oy} \boldsymbol{e}_{t} + \boldsymbol{W}_{oh} h_{t-1} + \boldsymbol{W}_{od} d_{t} + b_{o}\right)
\end{align}
With the signals for input gate, forget gate, and output gate, $h_{t}$ is computed as:
\begin{align}
c_{t} &= i_{t} \odot \phi\left(W_{cy} e_{t} + W_{ch} h_{t-1} + W_{cd} d_{t} + b_{c}\right) + f_{t} \odot c_{t-1} \\
h_{t} &= o_{t} \odot \tanh\left(c_{t}\right)
\end{align}
Both the context vector $d_{t}$ and $h_{t}$ are used to predict the next word $y_{t}$:
\begin{align}
\tilde{\boldsymbol{y}}_{t} &= \sigma\left(\boldsymbol{W}_{h} \boldsymbol{h}_{t} + \boldsymbol{W}_{d} \boldsymbol{d}_{t}\right) \\
y_{t} &\sim \operatorname{softmax}\left(\tilde{\boldsymbol{y}}_{t}\right)
\end{align}
This multimodal approach ensures a comprehensive understanding and integration of both visual and textual elements within scientific charts, ultimately improving the accuracy and relevance of generated titles. Future work will explore expanding the dataset, refining modal interaction mechanisms, and enhancing model robustness through advanced techniques like adversarial training and ensemble learning.

\section{Experiments}
\begin{table}[!t]  
  \centering  
  \begin{tabular}{cc}  
    \toprule  
    \textbf{Model} & \textbf{Test BLEU-4} \\ \midrule 
    TextCNN+LSTM & 0.0950 \\ 
    ResNet+LSTM+att & 0.0370 \\ 
    TextCNN+ResNet+LSTM+att & 0.1200 \\ 
    Transformer V2 & 0.0750 \\ 
    ClipCap & 0.0980 \\  \
    CLIP+RoBerta+GPT-2 & 0.1380 \\ 
    Our Model & 0.2050 \\ \bottomrule
  \end{tabular} 
  \caption{BLEU-4 Scores for Different Models}  
  \label{tab:main_results}
\end{table}
\subsection{Evaluation Metrics}
In this work, we use BLEU-4 as the evaluation metric, consistent with the benchmark model paper. BLEU, or Bilingual Evaluation Understudy, assesses the quality of machine translation by comparing machine-generated translations with human references, focusing on the precision of n-grams. BLEU-4, specifically, evaluates 4th-order n-grams (sequences of four consecutive words) to provide a detailed assessment of translation fluency and quality. This metric, widely used in machine translation, image captioning, text summarization, and chart title generation, measures both fluency and accuracy. The BLEU-4 score ranges from 0 to 1, with higher scores indicating closer alignment with human references, thus higher quality. For this task, BLEU-4 serves as a comprehensive metric for evaluating the generated chart titles at the sentence level.

\begin{table*}[!t]\small
  \centering  
  \begin{tabular}{cccccc}  
    \toprule  
    \textbf{ImageEncoder (ResNet)} & \textbf{ImageEncoder (CLIP)} & \textbf{TextEncoder} & \textbf{Co-Transformer} & \textbf{Add Related Knowledge} & \textbf{BLEU-4} \\ \midrule 
    \checkmark & & & & & 0.0370 \\  
    & & \checkmark & & & 0.0950 \\ 
    \checkmark & & \checkmark & \checkmark & & 0.1700 \\
    & \checkmark & \checkmark & \checkmark & & 0.2050 \\ 
    & \checkmark & \checkmark & \checkmark & \checkmark & 0.2150 \\ \bottomrule 
  \end{tabular} 
  \caption{BLEU-4 Scores for Ablation Studies}  
  \label{tab:ablation}
\end{table*}

\subsection{Experimental Setting}
We adopted a differential learning rate approach to fine-tune the pre-trained model for scientific chart title generation. Different optimization parameters were applied to various modules of the model. Specifically, the RoBERTa (base version from Huggingface) and ResNet (ResNet-101 version) models were used. The learning rates were set to 5e-5 for the modal interaction module, 1e-4 for the text and image encoders, and 5e-4 for the decoder. A weight decay coefficient of 1e-5 was applied. To prevent overfitting, an exponential learning rate decay strategy with a decay coefficient of 0.9 was implemented, along with gradient clipping to prevent gradient explosions.

\subsection{Main Results}
To evaluate the feasibility and challenges of creating an image-captioning model for scientific figures, we established several baselines and tested them using the SCICAP dataset. Caption quality was measured by BLEU-4, using the test set of the corresponding data collection as a reference.

The paper SCICAP serves as a crucial benchmark, presenting a dataset and designing three baseline models: Vision-only, Text-only, and Vision+Text.

(1) Vision-only Model: Utilizes ResNet for feature extraction from chart images, generating titles through a recurrent neural network, focusing on the image modality.
(2) Text-only Model: Generates captions solely based on chart-related texts, using TextCNN and an attention-based Long Short-Term Memory network.
(3) Vision+Text Model: Fuses text and visual information using a two-tower approach. ResNet extracts features from chart images, and TextCNN processes related texts. These features are concatenated and input into a decoder for caption generation.

Additionally, Chart2Text encodes chart-related text information using a Transformer encoder and generates captions through a Transformer decoder. We also compared our method with ClipCap, which uses the pre-trained CLIP model to encode image information and a Transformer decoder for caption generation.

Our multimodal scientific chart caption generation model outperformed all baseline models on the SCICAP dataset, as detailed in Table \ref{tab:main_results}, demonstrating the effectiveness of our approach.

\subsection{Ablation Study}
Ablation experiments were conducted to verify the effectiveness of each module. The proposed multimodal scientific chart caption generation method was compared with unimodal methods and a method utilizing both text and image information without the modal interaction layer. These comparisons highlight the contribution and effectiveness of each module in the caption generation process.

Results from the ablation studies, shown in Table \ref{tab:ablation}, demonstrate that the multimodal approach significantly outperforms unimodal methods. Specifically, the multimodal method improves performance by 8 percentage points over text-only methods and 12 percentage points over image-only methods. Compared to a two-tower model for modal fusion and interaction, our method further enhances performance by 5 percentage points, underscoring the effectiveness of our approach.

\section{Conclusion and Future Work}

This paper presents a novel multimodal chart caption generation scheme, addressing the limitations of previous monomodal approaches. Our proposed method significantly improves both accuracy and fluency in generating chart titles by utilizing a multimodal encoding and co-attention mechanism within an encoder-decoder framework. Implemented using the PyTorch framework, our model employs pre-trained RoBERTa and ResNet for encoding text and image information, respectively, and leverages parameter initialization to accelerate model convergence. Detailed descriptions of the experimental parameters, hardware, and training processes are provided, demonstrating the effectiveness of our approach through comprehensive analysis and experimental results.

Despite the advancements made, several research directions remain open for further exploration:

\begin{enumerate}
    \item There is a scarcity of datasets specifically targeting scientific charts. Constructing a more segmented chart title dataset, incorporating contextual information about the charts, can significantly advance this field.
    \item The current co-attention mechanism encodes each modality separately, interacting only through self-attention. Addressing modal alignment issues, possibly by adopting an early fusion approach, could enable more effective interaction within the same feature encoder.
    \item Leveraging techniques such as adversarial networks, ensemble learning, and others can improve the robustness and generalization capability of the title generation model, making it more adaptable to various domains and scenarios.
\end{enumerate}

In conclusion, our work provides a solid foundation for the automatic generation of scientific chart titles, integrating multimodal data to produce accurate and contextually relevant titles. Continued research in the outlined directions will further enhance the quality and applicability of automated chart title generation systems.

\bibliographystyle{IEEEbib}
\bibliography{ref}

\begin{thebibliography}{10}

\bibitem{Zhang2023}
Zhaoyi Sun, Mingquan Lin, Qingqing Zhu, Qianqian Xie, Fei Wang, Zhiyong Lu, and Yifan Peng,
\newblock ``A scoping review on multimodal deep learning in biomedical images and texts,''
\newblock {\em Journal of Biomedical Informatics}, p. 104482, 2023.

\bibitem{zhou2024visual}
Yucheng Zhou, Xiang Li, Qianning Wang, and Jianbing Shen,
\newblock ``Visual in-context learning for large vision-language models,''
\newblock {\em arXiv preprint arXiv:2402.11574}, 2024.

\bibitem{zhou2022claret}
Yucheng Zhou, Tao Shen, Xiubo Geng, Guodong Long, and Daxin Jiang,
\newblock ``Claret: Pre-training a correlation-aware context-to-event transformer for event-centric generation and classification,''
\newblock in {\em Proceedings of the 60th Annual Meeting of the Association for Computational Linguistics (Volume 1: Long Papers)}, 2022, pp. 2559--2575.

\bibitem{Baltrusaitis2018}
Tadas Baltru{\v{s}}aitis, Chaitanya Ahuja, and Louis-Philippe Morency,
\newblock ``Multimodal machine learning: A survey and taxonomy,''
\newblock {\em IEEE transactions on pattern analysis and machine intelligence}, vol. 41, no. 2, pp. 423--443, 2018.

\bibitem{Arevalo2017}
J.~Arevalo, T.~Solorio, M.~Montes-y Gomez, and F.A. Gonzalez,
\newblock ``Gated multimodal units for information fusion,''
\newblock in {\em 5th International Conference on Learning Representations}, 2017.

\bibitem{Li2018}
Yingming Li, Ming Yang, and Zhongfei Zhang,
\newblock ``A survey of multi-view representation learning,''
\newblock {\em IEEE transactions on knowledge and data engineering}, vol. 31, no. 10, pp. 1863--1883, 2018.

\bibitem{Ramachandram2017}
Dhanesh Ramachandram and Graham~W Taylor,
\newblock ``Deep multimodal learning: A survey on recent advances and trends,''
\newblock {\em IEEE signal processing magazine}, vol. 34, no. 6, pp. 96--108, 2017.

\bibitem{Mogadala2019}
Aditya Mogadala, Marimuthu Kalimuthu, and Dietrich Klakow,
\newblock ``Trends in integration of vision and language research: A survey of tasks, datasets, and methods,''
\newblock {\em Journal of Artificial Intelligence Research}, vol. 71, pp. 1183--1317, 2021.

\bibitem{Gao2020}
Jing Gao, Peng Li, Zhikui Chen, and Jianing Zhang,
\newblock ``A survey on deep learning for multimodal data fusion,''
\newblock {\em Neural Computation}, vol. 32, no. 5, pp. 829--864, 2020.

\bibitem{Radford2021}
Alec Radford, Jong~Wook Kim, Chris Hallacy, Aditya Ramesh, Gabriel Goh, Sandhini Agarwal, Girish Sastry, Amanda Askell, Pamela Mishkin, Jack Clark, et~al.,
\newblock ``Learning transferable visual models from natural language supervision,''
\newblock pp. 8748--8763, 2021.

\bibitem{zhou2021improving}
Yucheng Zhou, Xiubo Geng, Tao Shen, Wenqiang Zhang, and Daxin Jiang,
\newblock ``Improving zero-shot cross-lingual transfer for multilingual question answering over knowledge graph,''
\newblock in {\em Proceedings of the 2021 Conference of the North American Chapter of the Association for Computational Linguistics: Human Language Technologies}, 2021, pp. 5822--5834.

\bibitem{zhou2021modeling}
Yucheng Zhou, Xiubo Geng, Tao Shen, Jian Pei, Wenqiang Zhang, and Daxin Jiang,
\newblock ``Modeling event-pair relations in external knowledge graphs for script reasoning,''
\newblock {\em Findings of the Association for Computational Linguistics: ACL-IJCNLP 2021}, 2021.

\bibitem{wang2024memorymamba}
Qianning Wang, He~Hu, and Yucheng Zhou,
\newblock ``Memorymamba: Memory-augmented state space model for defect recognition,''
\newblock {\em arXiv preprint arXiv:2405.03673}, 2024.

\bibitem{Liu2023SciCap}
Zhishen Yang, Raj Dabre, Hideki Tanaka, and Naoaki Okazaki,
\newblock ``Scicap+: A knowledge augmented dataset to study the challenges of scientific figure captioning,''
\newblock {\em arXiv preprint arXiv:2306.03491}, 2023.

\bibitem{zhou2022eventbert}
Yucheng Zhou, Xiubo Geng, Tao Shen, Guodong Long, and Daxin Jiang,
\newblock ``Eventbert: A pre-trained model for event correlation reasoning,''
\newblock in {\em Proceedings of the ACM Web Conference 2022}, 2022, pp. 850--859.

\bibitem{Xu2015ShowAT}
Kelvin Xu, Jimmy Ba, Ryan Kiros, Kyunghyun Cho, Aaron Courville, Ruslan Salakhutdinov, Richard Zemel, and Yoshua Bengio,
\newblock ``Show, attend and tell: Neural image caption generation with visual attention,''
\newblock in {\em International conference on machine learning}. PMLR, 2015, pp. 2048--2057.

\bibitem{You2016ImageCW}
Quanzeng You, Hailin Jin, Zhaowen Wang, Chen Fang, and Jiebo Luo,
\newblock ``Image captioning with semantic attention,''
\newblock in {\em Proceedings of the IEEE conference on computer vision and pattern recognition}, 2016, pp. 4651--4659.

\bibitem{Vinyals2015ShowAT}
Oriol Vinyals, Alexander Toshev, Samy Bengio, and Dumitru Erhan,
\newblock ``Show and tell: A neural image caption generator,''
\newblock in {\em Proceedings of the IEEE conference on computer vision and pattern recognition}. IEEE, 2015, pp. 3156--3164.

\bibitem{Mokady2021ClipCap}
Ron Mokady, Amir Hertz, and Amit~H Bermano,
\newblock ``Clipcap: Clip prefix for image captioning,''
\newblock {\em arXiv preprint arXiv:2111.09734}, 2021.

\bibitem{Radford2021LearningTV}
Alec Radford, Jong~Wook Kim, Chris Hallacy, Aditya Ramesh, Gabriel Goh, Sandhini Agarwal, Girish Sastry, Amanda Askell, Pamela Mishkin, Jack Clark, et~al.,
\newblock ``Learning transferable visual models from natural language supervision,''
\newblock {\em arXiv preprint arXiv:2103.00020}, 2021.

\bibitem{Zhang2023CapText}
Shinjini Ghosh and Sagnik Anupam,
\newblock ``Captext: Large language model-based caption generation from image context and description,''
\newblock {\em arXiv preprint arXiv:2306.00301}, 2023.

\bibitem{Wang2019FigureCR}
Charles Chen, Ruiyi Zhang, Eunyee Koh, Sungchul Kim, Scott Cohen, Tong Yu, Ryan Rossi, and Razvan Bunescu,
\newblock ``Figure captioning with reasoning and sequence-level training,''
\newblock {\em arXiv preprint arXiv:1906.02850}, 2019.

\bibitem{Jin2023SummariesAC}
Chieh-Yang Huang, Ting-Yao Hsu, Ryan Rossi, Ani Nenkova, Sungchul Kim, Gromit Yeuk-Yin Chan, Eunyee Koh, Clyde~Lee Giles, and Ting-Hao'Kenneth' Huang,
\newblock ``Summaries as captions: Generating figure captions for scientific documents with automated text summarization,''
\newblock {\em arXiv preprint arXiv:2302.12324}, 2023.

\bibitem{Lee2023FigCapsHF}
Ashish Singh, Prateek Agarwal, Zixuan Huang, Arpita Singh, Tong Yu, Sungchul Kim, Victor Bursztyn, Nikos Vlassis, and Ryan~A Rossi,
\newblock ``Figcaps-hf: A figure-to-caption generative framework and benchmark with human feedback,''
\newblock {\em arXiv preprint arXiv:2307.10867}, 2023.

\bibitem{ICCV2023}
Dian Chao, Xin Song, Shupeng Zhong, Boyuan Wang, Xiangyu Wu, Chen Zhu, and Yang Yang,
\newblock ``The solution for the iccv 2023 1st scientific figure captioning challenge,''
\newblock {\em arXiv preprint arXiv:2403.17342}, 2024.

\bibitem{Zhang2023SciCapenter}
Ting-Yao Hsu, Chieh-Yang Huang, Shih-Hong Huang, Ryan Rossi, Sungchul Kim, Tong Yu, C~Lee Giles, and Ting-Hao~Kenneth Huang,
\newblock ``Scicapenter: Supporting caption composition for scientific figures with machine-generated captions and ratings,''
\newblock pp. 1--9, 2024.

\end{thebibliography}
\end{document}